# Can reasoning models comprehend mathematical problems in Chinese ancient texts? An empirical study based on data from Suanjing Shishu

Chang Liu[1], Dongbo Wang[1], Liu liu[1], Zhixiao Zhao[1]

1 (College of Information Management, Nanjing Agricultural University, Nanjing, China)

**Abstract:** This study addresses the challenges in intelligent processing of Chinese ancient mathematical classics by constructing Guji_MATH, a benchmark for evaluating classical texts based on Suanjing Shishu (《算经十书》). It systematically assesses the mathematical problem-solving capabilities of mainstream reasoning models under the unique linguistic constraints of classical Chinese. Through machine-assisted annotation and manual verification, 538 mathematical problems were extracted from 8 canonical texts, forming a structured dataset centered on the "Question-Answer-Solution" framework, supplemented by problem types and difficulty levels. Dual evaluation modes—closed-book (autonomous problem-solving) and open-book (reproducing classical solution methods)—were designed to evaluate the performance of six reasoning models on ancient Chinese mathematical problems. Results indicate that reasoning models can partially comprehend and solve these problems, yet their overall performance remains inferior to benchmarks on modern mathematical tasks. Enhancing models' classical Chinese comprehension and cultural knowledge should be prioritized for optimization. This study provides methodological support for mining mathematical knowledge from ancient texts and disseminating traditional culture, while offering new perspectives for evaluating cross-linguistic and cross-cultural capabilities of reasoning models.

## 0 Introduction

As one of the earliest civilizations to develop mathematics, China accumulated a vast repository of mathematical knowledge over its long history. From the emergence of numerical symbols and the heavenly stems and earthly branches system in ancient times, to the birth of mature mathematical frameworks in the Qin and Han dynasties, followed by the flourishing of mathematical theories during the Song and Yuan periods, and the integration of Eastern and Western mathematical practices in the Ming and Qing dynasties, ancient Chinese mathematics consistently exhibited a unique interplay of practical application and theoretical exploration. Effectively leveraging this critical cultural heritage—ancient mathematical texts—is essential for understanding China's historical mathematical wisdom and fostering international exchange of classical mathematical ideas. However, due to the inherent complexity of classical Chinese and the distinct computational methods employed in ancient mathematics, modern scholars struggle to interpret these texts, risking the loss of traditional calculation logic and techniques.

In recent years, rapidly advancing large language models (LLMs) and LLM-based reasoning models have demonstrated exceptional performance in tasks requiring complex reasoning, such as mathematical computation and code generation. These advancements offer new momentum for intelligent processing of ancient texts. Yet, existing research on AI-driven classical text analysis predominantly focuses on literary and historical documents, with limited attention to solving mathematically oriented problems expressed in classical Chinese—an interdisciplinary domain integrating humanities and STEM. Furthermore, current benchmarks for evaluating mathematical reasoning in LLMs are almost exclusively built on English texts, leaving a gap in understanding how linguistic characteristics influence model performance. To address these limitations, this study pioneers the use of mathematical problems from Suanjing Shishu (《算经十书》), the most renowned collection of Chinese mathematical works, to construct a reasoning model evaluation benchmark. We systematically evaluate and analyze the performance of mainstream reasoning models on ancient mathematical tasks. Through empirical investigation, this study aims to address three key questions:

Q1: How can the "Question-Answer-Solution" structure in Chinese mathematical classics be utilized to formulate problems that enable reasoning models to think, respond, and verify solutions?

Q2: How can the Guji_MATH benchmark assess the accuracy and efficiency of general-purpose reasoning models in solving ancient mathematical problems?

Q3: How can an analytical framework be constructed to identify mathematical problem types that reasoning models excel at and quantitatively diagnose error causes?

Using machine-assisted annotation, we developed the Guji_MATH benchmark comprising 538 mathematical problems. We deployed mainstream reasoning models to solve these problems and analyzed the experimental results. By systematically addressing these questions, this study reveals the potential and limitations of large-scale reasoning models in tackling ancient mathematical challenges—a domain bridging humanities and technical disciplines. It provides a methodological foundation for AI-driven knowledge mining from classical mathematical texts and introduces a novel perspective for evaluating mathematical reasoning capabilities: assessing models through problems described in non-dominant languages. This approach also offers insights into how linguistic features influence mathematical reasoning.

The dataset, code, and experimental results generated in this study have been open-sourced. (https://github.com/Xunzi-LLM-of-Chinese-classics/Guji_Math)

# 1 Related works

## 1.1 Large language model and Reasoning model

Large language models (LLMs) refer to language models based on deep neural network architectures, possessing billions or more learnable parameters, and trained on massive text corpora (W. X. Zhao et al., 2025). Early exploratory LLMs such as T5 (Raffel et al., 2020) and GPT-3 (Brown et al., 2020) were built upon the Transformer architecture (Vaswani et al., 2017). During this period, standardized training paradigms had not yet been established, and researchers predominantly relied on empirical scaling laws (Kaplan et al., 2020; Hoffmann et al., 2022) to progressively increase model parameters and training dataset sizes. Through the combination of pre-training, fine-tuning, and prompt learning (Liu et al., 2023), these models acquired language processing capabilities and world knowledge, enabling them to understand user queries and generate appropriate responses. However, LLMs from this era still struggled with solving complex problems. In late 2022, OpenAI released ChatGPT (ChatGPT, 2022), which demonstrated substantially enhanced linguistic capabilities and knowledge utilization compared to previous models, marking a milestone in LLM development. While generating significant societal impact, the construction, capability evaluation, and domain applications of LLMs gradually attracted widespread research attention. Generally, ChatGPT-style LLMs established stable and standardized training processes, dividing LLM development into three key stages:

(1) Pre-training: Training models on massive unannotated text data

(2) Instruction tuning (J. Wei, Bosma, et al., 2022): Enhancing models' ability to properly respond to user instructions through supervised training on human-annotated instruction-response pairs

(3) Reinforcement Learning from Human Feedback (RLHF) (Ouyang et al., 2022): Further optimizing SFT (Supervised Fine-tuning) models using reinforcement learning algorithms with preference data to improve response safety, usefulness, and robustness

This three-stage optimization framework significantly enhanced LLMs' question-answering performance. Numerous institutions subsequently adopted this approach to develop and release their own LLMs, including Llama (Touvron et al., 2023), Qwen (Bai et al., 2023), and GLM (GLM et al., 2024), providing high-quality linguistic infrastructure for open-source communities. Through post-training, retrieval-augmented generation (RAG) (P. Zhao et al., 2024), and advanced AI Agent (Xi et al., 2025) development, LLMs can now deeply integrate with both general and vertical industry scenarios to solve problems beyond their native capabilities.

Traditional large language models (LLMs) have achieved success in many fields, yet they still exhibit significant limitations in scenarios requiring complex logical reasoning and multi-step problem-solving. To address these challenges, researchers have begun exploring specialized training approaches using mathematical and code datasets to enhance step-by-step problem-solving capabilities. The CoT (chain-of-thought) method (J. Wei, Wang et al., 2022) has emerged as a key technique to extend models' reasoning functions by forcing them to decompose complex tasks into intermediate steps for gradual deduction. While this approach effectively improves LLMs' performance on complex reasoning tasks (Sprague et al., 2024), the shallow and linear nature of short reasoning chains limits models' ability to explore alternative solution paths and makes them susceptible to error propagation. Consequently, models employing short reasoning chains still struggle to simulate sophisticated human cognitive processes and achieve expert-level performance in testing scenarios.

Researchers are now investigating tree-structured reasoning processes (both explicit and implicit) to encourage models to explore more beneficial solution paths during reasoning while reflecting on previous steps to optimize final outcomes. Representative achievements in this direction include OpenAI's O1 model ("Learning to Reason with LLMs", n.d.) and Deepseek's R1 model (DeepSeek-AI et al., 2025), which demonstrate that long-chain reasoning models can achieve human-expert-level performance across various challenging benchmarks. Recent studies (Chen et al., 2025) have conducted in-depth analyses of long-chain reasoning phenomena in LLMs, investigating aspects such as the emergence of extended reasoning chains, test-time reasoning extensions, and overthinking patterns. These investigations aim to refine existing reasoning models through theoretical insights and evaluate their performance across general and domain-specific evaluation benchmarks.

## 1.2 Benchmark for Evaluating the Mathematical Capabilities of Large Language Models

The ability to accurately and efficiently understand and solve mathematical problems is a critical metric for assessing the reasoning capabilities of large language models and reasoning models. Current mathematical evaluation benchmarks primarily consist of two types of tasks: mathematical problem-solving and theorem proving (Ahn et al., 2024). Within mathematical problem-solving tasks, subcategories include arithmetic problems, mathematical word problems, geometry problems, and math problems in visual-language contexts. Typically, these benchmarks are constructed by manually or automatically annotating existing math problems, including their questions, reasoning steps, and answers. Large models are then prompted to generate responses, from which answers are extracted and evaluated for accuracy. Existing benchmarks for evaluating unimodal large language models and reasoning models predominantly focus on mathematical word problems, where problems are described in natural language rather than symbolic notation, requiring models to comprehend the problem's context and provide solutions. Notable examples include MATH (Hendrycks et al., 2021), which compiles 12,500 math problems from U.S. high school competitions, each accompanied by complete solution steps, final answers, and difficulty ratings. GSM8K (Cobbe et al., 2021) collects 8.5K elementary-level math problems created by human writers, with each problem requiring 2–8 steps involving basic arithmetic operations. PRM800K divides the MATH dataset into a 12K-problem training set and a 500-problem test set. The training set includes 75K generated solutions and 800K step-level annotations, forming the basis for training process- and result-based reward models. The remaining 500 test problems are known as the MATH-500 benchmark (HuggingFaceH4/MATH-500 · Datasets at Hugging Face, 2025), widely used recently to evaluate reasoning models. The AIME benchmark (HuggingFaceH4/Aime_2024 · Datasets at Hugging Face, 2025) aggregates annual problems from the U.S. Mathematical Invitational Examination. Despite its limited problem count, its high difficulty and timeliness have garnered significant institutional attention, establishing it as a key benchmark for assessing reasoning capabilities.

In Chinese mathematical benchmarks, Cmath (T. Wei et al., 2023) compiles 1,700 math problems from primary school exercises and exams, each annotated with correct answers to evaluate models' understanding of Chinese math problems. TAL-SCQ5K (Math-Eval/TAL-SCQ5K · Datasets at Hugging Face, n.d.) gathers 5K Chinese-English math problems from competitions across grade levels, translated into English to train and evaluate large language models. GAOKAO-Bench (Zhang et al., 2024) compiles questions from China's national college entrance examination (Gaokao), covering both subjective and objective math problems. As China's most standardized, comprehensive, and widely recognized exam, Gaokao-based benchmarks effectively assess models' logical reasoning abilities.

Overall, existing mathematical evaluation benchmarks exhibit the following notable characteristics:

(1) Most benchmarks prioritize modern math problems, drawing from English-language standardized tests or competitions. Their phrasing and problem-solving logic reflect contemporary mathematical education systems.

(2) While Chinese benchmarks have achieved progress, they predominantly cover conventional problems from basic education. They lack systematic coverage of ancient math problems expressed in classical Chinese and traditional solution methods.

(3) Research on cross-linguistic mathematical reasoning remains insufficient, particularly regarding how non-Latin scripts (e.g., classical Chinese) affect model performance. This oversight risks cultural bias in current evaluation frameworks, limiting their ability to fully capture models' math comprehension and application across diverse cultural and communicative contexts.

## 1.3 The Ten Mathematical Classics and the Compilation of Ancient Chinese Mathematical Texts

Throughout the long history of struggle against nature and social practice, ancient Chinese people accumulated rich mathematical experiences and established a unique, practice-oriented mathematical system. However, China's historical decline in the modern era limited academic exchanges between East and West, leading to widespread neglect of ancient Chinese mathematical achievements. Starting in the 1950s, the renowned sinologist Joseph Needham systematically compiled *Science and Civilisation in China* (Needham, J., 1959) based on extensive research, introducing a series of ancient Chinese scientific and technological achievements—including mathematics—to the West. This work overturned prevailing misconceptions about China's mathematical heritage. In Volume 3 of the series, Dr. Needham cataloged mathematical literature and accomplishments from prehistoric times to the Qing Dynasty, dividing Chinese mathematical history into three periods: from antiquity to the Three Kingdoms period, the Three Kingdoms to the early Song Dynasty, and the Song-Yuan-Ming eras, corresponding respectively to the origin, development, and maturation of the Chinese mathematical system. *Suanjing Shishu* used in this study originate from the first two periods. Compiled and annotated by Tang Dynasty scholars, these texts became textbooks for the mathematics department of National Imperial Academy, representing the most significant works reflecting ancient Chinese mathematical thought. Among these, *The Nine Chapters on the Mathematical Art* (《九章算术》),the earliest mathematical text compiled over 2,000 years ago, stands as the cornerstone of the Chinese mathematical system. Its problems are structured in a "Question-Answer-Method" format, covering nine categories: *Fangtian* (land area calculations), *Sumi* (commodity trade calculations), *Cuifen* (allocation algorithms), *Shaoguang (Square Root Extraction)*, *Shanggong* (construction-related problems), *Junshu（Taxation Calculation）*, *Yingbuzu （Excess-Deficit Method）*, *Fangcheng (* Linear Equations), and *Gougu (Pythagorean Theorem Applications)*. The text pioneered global advancements in fractional systems, negative numbers, linear equation solutions, and the Pythagorean theorem.Contemporaneously, *The Zhou Bi Suan Jing* （《周髀算经》） integrated astronomical calculations with mathematics, providing the first proof of the Pythagorean theorem. *The Sea Island Mathematical Classic （《海岛算经》）*, originally a continuation of the Pythagorean chapter in *The Nine Chapters* by the Three Kingdoms mathematician Liu Hui, describes nine problems on measuring heights and depths using repeated observations and difference calculations with gnomon tools. Texts like *Sunzi Suanjing (《孙子算经》）, Zhang Qiujian Suanjing （《张邱建算经》）, Xiahou Yang Suanjing （《夏侯阳算经》）*, and *Wucao Suanjing* （《五曹算术》）emerged during the Wei-Jin-Southern and Northern Dynasties, documenting cutting-edge mathematical techniques applied to area, volume, and quantity calculations. *Wujing Suanshu (《五经算术》)*, closely tied to Confucian studies, provided annotations on mathematical problems in classical texts, offering value for Confucian scholarship. The final text, *Jigu Suanjing (《缉古算经》）*, written in the Tang Dynasty, addressed complex problems in volume calculation, Pythagorean geometry, cubic equations, and biquadratic equations. These solutions were among the most advanced globally at the time. *Zhuishu* （《缀术》） and *Shushu Jiyi (《数术记遗》）* are unique within the collection: the former, authored by the renowned mathematician Zu Chongzhi, contained profound mathematical theories but has since been lost. The latter records 14 ancient Chinese calculation methods and tools, 13 of which are now extinct. Modern scholars have attempted to reconstruct these lost techniques. Since *Zhuishu* is irretrievable, this study adopts the 1963 edition edited by Qian Baocong, substituting *Shushu Jiyi* for *Zhuishu* to align with the standard

*Ten Mathematical Classics.*

In recent decades, historians of mathematics have established a modern interpretive framework for ancient Chinese mathematics through textual collation, annotation, and cross-cultural comparative studies. For example, Qian Baocong (QIAN, B. C., 1963) and Guo Shuchun (Guo, S. C. & Liu, D., 1998) systematically annotated variants in different editions of the *Suanjing Shishu*, revealing the original forms of high-quality ancient texts to the public. In the era of rapid information technology and AI development, researchers have explored digitization and intelligent processing of classical texts. For instance, the 2024 academic symposium in Inner Mongolia (Yang et al., 2024) discussed innovative digital humanities approaches to organizing and promoting mathematical classics, employing technologies like image recognition, knowledge graph construction, and LLM-based agents to automatically structure knowledge based on the unique features of ancient mathematical texts, achieving promising results. This study similarly leverages state-of-the-art AI technologies, but from an evaluation perspective, to explore reasoning models' comprehension of ancient Chinese mathematical texts.

## 2 Methods

The Research Framework of This Study is Illustrated in Figure 1:

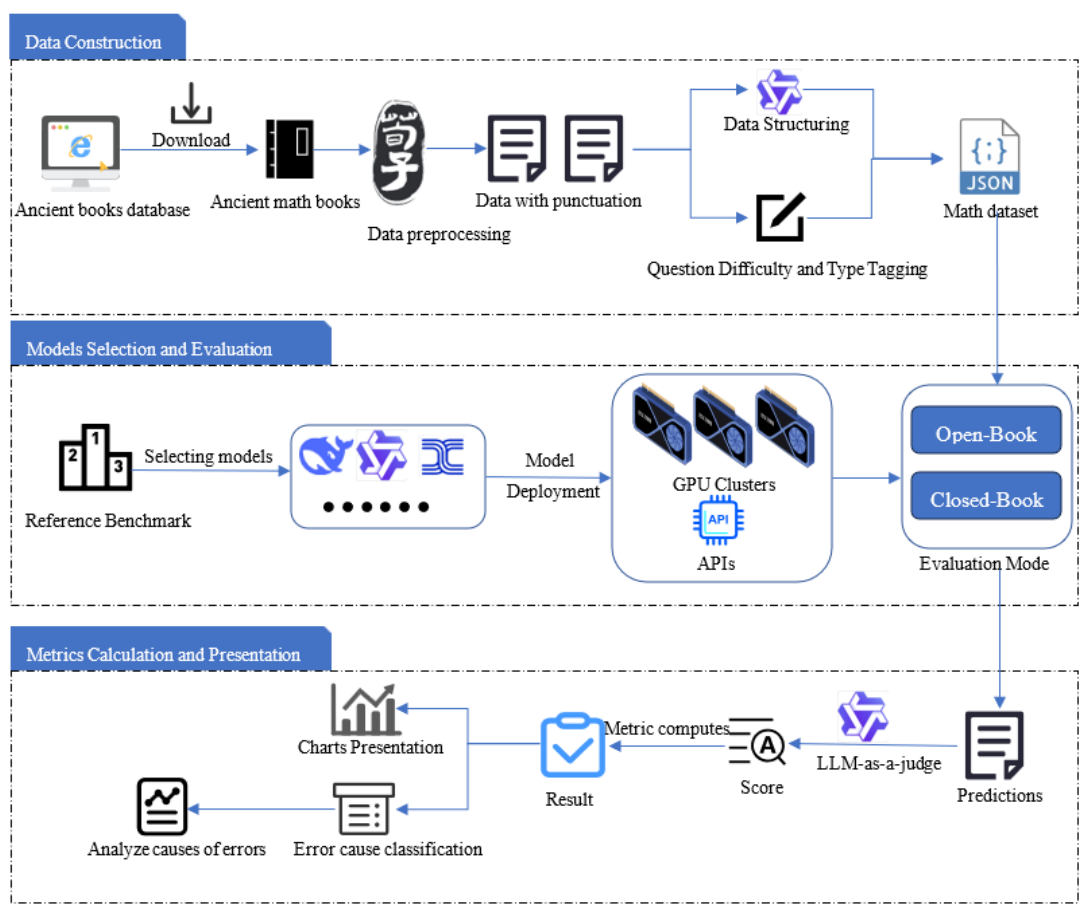

**Figure 1.** Research Framework

The research framework of this study is illustrated in Figure 1. This framework encompasses three sequential phases:

(1) The construction of an ancient mathematical problem dataset, the testing of reasoning models, and the analysis of results: In the dataset construction phase, high-quality texts from *Suanjing Shishu* are first collected. The Xunzi series of models are employed to annotate punctuation marks, enabling the extraction of all problem descriptions, standard answers, and solution steps. Following existing protocols for mathematical benchmark construction, the extracted content is organized into a "Question-Answer-Solution" triplet structure. Subsequently, manual review is conducted to correct errors and supplement missing conditions, while a hybrid approach combining machine and human efforts assigns difficulty levels and mathematical methods to each problem. The final processed dataset is saved in JSON format for evaluation of reasoning models.

(2) Reasoning model testing phase: Chinese closed-source reasoning models are accessed via API, while open-source reasoning models are deployed offline. Two types of prompts are used to guide the models in answering ancient mathematical problems.

(3) Result analysis phase: an LLM is utilized as a judge by inputting the models' predictions and ground-truth answers to compute accuracy metrics. This enables an analysis of how different reasoning models perform on mathematical problems, with further exploration of which types of problems from classical texts each model is better suited to handle, based on problem difficulty and category. Finally, manually annotated and quantitative analysis of incorrectly answered problems is conducted to identify factors hindering the models' ability to solve ancient mathematical problems.

Through these steps, we constructed the classical Chinese mathematical evaluation benchmark *Guji_MATH*, which provides a comprehensive understanding of how reasoning models perform on mathematical problems and explores the feasibility of applying such models to interpret ancient mathematical texts. All prompts used to invoke large language models for data processing mentioned in this section are included in the appendix of this article.

## 2.1 Data Construction

### 2.1.1 Data collection and preprocessing

This study first collected the complete text of *Suanjing Shishu* as the corpus source. To obtain high-quality data, we downloaded the contents of the ten mathematical classics from the official website of the Chinese Philosophy E-Book Project (Sturgeon, n.d.). On this platform, *The Nine Chapters on the Mathematical Art*, *Sunzi suanjing*, *The Sea Island Mathematical Classic*, and *Zhou Bi Suan Jing* had already undergone manual collation and punctuation annotation, making their texts directly usable. The remaining six texts lacked syntactic-level punctuation markers. Figure 2 displays the content of *Xiahouyang Suanjing* from the Ctext platform:

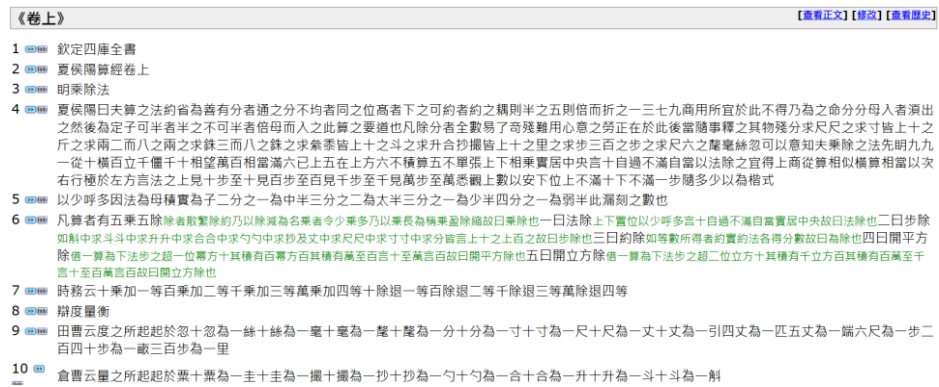

**Figure 2.** *Xiahouyang Suanjing* on Ctext

As shown in Figure 2, the texts of these six remaining mathematical classics originate from the digitized version of the *Siku Quanshu* database. After OCR (Optical Character Recognition), the resulting text is segmented and displayed on the website frontend. Since punctuation-free text hinders the model's ability to understand and segment meaningful structures, this study employed the Xunzi series dialogue models to add appropriate punctuation. The Xunzi series large language models, developed and open-sourced by a team from Nanjing Agricultural University (Xunzi-LLM-of-Chinese-Classics/XunziALLM, 2023/2025), are specialized for intelligent processing of classical texts and demonstrate strong performance in tasks such as text translation, punctuation annotation, and entity labeling. In this study, we used the recommended punctuation task prompt from the Xunzi model in combination with unpunctuated text, treating each paragraph as a processing unit to generate punctuation results.

## 2.1.2 Extraction and Structuring of Mathematical Problems from Classical Texts

The typical structure of mathematical problems in *Suanjing Shishu* follows a "Question-Answer-Solution" ("问—答—术") format, where problem descriptions, standard answers, and corresponding solution methods appear adjacently and consecutively in the text. While this structure provides a natural foundation for automated extraction, simply extracting adjacent triplets of text often leads to incomplete problem conditions, reducing the accuracy of model responses. For example, in the *Sumi* chapter of *The Nine Chapters on the Mathematical Art*, the author initially outlines grain conversion ratios, but subsequent problems and answers do not restate this critical premise. Additionally, problems within the same section may share implicit assumptions mentioned only in the first question. For instance, the fourth and fifth problems in the "Weight Measurement" section of *Xiahouyang Suanjing* share the same premise: during smelting, every jin ("斤",Chinese pound) of yellow iron reduces by three liang ("两",Chinese ounces). However, the fifth problem's description omits this premise, merely referencing "according to the previously mentioned consumption rate." Extracting such problems without considering shared or implicit premises inevitably results in missing conditions.

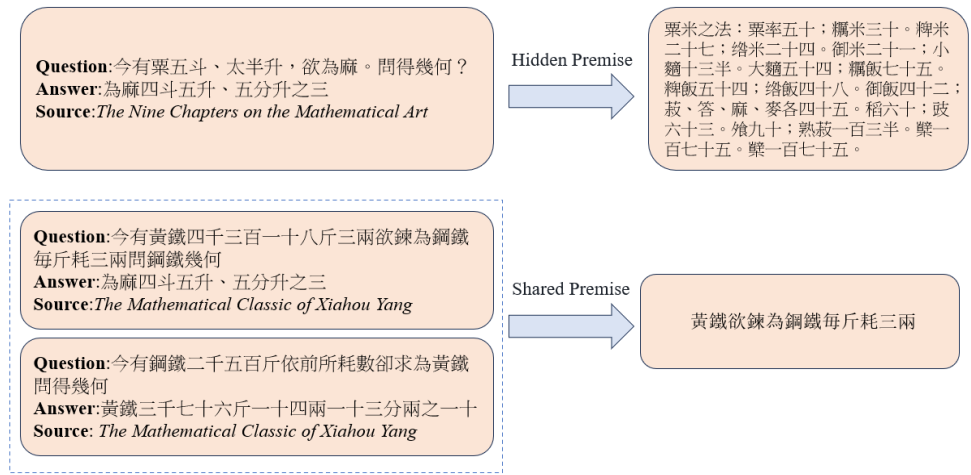

**Figure 3.** Utilizing Default Premise Conditions in Mathematical Problems from Classical Texts

**Figure 3** illustrates two scenarios of implicit and shared premise conditions in classical mathematical texts. In *The Nine Chapters on the Mathematical Art*, a problem involves converting millet into hemp and calculating the resulting quantity of hemp. However, the original text only includes the problem description and answer section, omitting the previously mentioned grain exchange ratio. In *Xiahouyang Suanjing*, two adjacent problems describe the conversion of yellow iron into steel during metallurgy. The premise condition appears in the fourth problem: when smelting yellow iron into steel, each jin (Chinese pound) reduces by three liang (Chinese ounces). Yet, the fifth problem's description omits this premise, merely referencing "according to the previously mentioned consumption rate." If such problems with default conditions are automatically extracted without contextual analysis, critical information will inevitably be lost.

To address this issue, we first employed the Qwen2.5 model (Qwen et al., 2025) to extract all "Question-Answer-Solution" ("问—答—术") triplet texts from the full corpus. After saving the results in JSON format and conducting manual verification, human annotators analyzed each text in conjunction with the full context to identify hidden or shared premises. These premises were then appended as new JSON attributes to the corresponding mathematical problem triplets, ensuring that every problem theoretically contained sufficient information for resolution by humans or machines. Through this process, we extracted 538 mathematical problems and their answers from the ten classics, including 511 problems with solution steps and 115 problems requiring explicitly noted premises. **Table 1** summarizes the number of extracted problems, solution methods, and premise conditions across the ten texts.

**Table 1** Statistical Results of Math Problem Extraction

| Classic Name | Extracted Problems | Proportion (%) | Solution Methods | Premise Conditions |
|---|---|---|---|---|
| *The Nine Chapters on the Mathematical Art* | 199 | 36.99% | 172 | 22 |
| *Zhou Bi Suan Jing* | 0 | 0.00% | 0 | 0 |
| *The Sea Island Mathematical Classic* | 9 | 1.67% | 9 | 0 |
| *Zhang Qiujian Suanjing* | 82 | 15.24% | 82 | 14 |
| *Xiahouyang Suanjing* | 80 | 14.87% | 80 | 67 |
| *Wujing Suanshu* | 19 | 3.53% | 19 | 0 |
| *Jigu Suanjing* | 20 | 3.72% | 20 | 0 |
| *Wucao Suanjing* | 66 | 12.27% | 66 | 0 |
| *Sunzi Suanjing* | 63 | 11.71% | 63 | 8 |
| *Shushu Jiyi* | 0 | 0.00% | 0 | 0 |

The extraction process and statistical results indicate that *Suanjing Shishu* do not uniformly contain mathematical problems, nor do all problems include corresponding solution methods. Since *The Zhou Bi Suan Jing* and *Shushu Jiyi* originally only present introductions to mathematical methods and proofs of mathematical theories—rather than fixed-answer arithmetic or word problems—we excluded these contents from the dataset. Among the remaining eight classics, a significant number of mathematical problems were extractable; however, 27 problems in the original texts lacked explicit solution steps. This was primarily because their solution methods closely resembled those of adjacent problems, and authors omitted repetitive explanations. Premise-dependent problems are concentrated in the *Sumi* chapter of *The Nine Chapters on the Mathematical Art* and in *Xiahouyang Suanjing*, with most premises being implicit and primarily related to unit conversion explanations.

**2.1.3 Problem Type and Difficulty Level Annotation**

To systematically analyze the performance differences of reasoning models across mathematical problems of varying complexity, this study combined characteristics of the ancient Chinese mathematical system with modern mathematical classification standards. A closed-source Qwen reasoning model with the largest parameter count was invoked to structurally annotate the 538 extracted mathematical problems along two dimensions: difficulty level and problem type.

**2.1.3.1 Question type Annotation**

Given the static nature of the dataset and the finite scope of mathematical theories and methods in ancient Chinese mathematical texts, these can be exhaustively enumerated. Joseph Needham's *Science and Civilisation in China* previously classified mathematical techniques and methods in ancient Chinese texts and provided examples corresponding to each category. This study referenced Needham's classification framework, categorizing the mathematical methods reflected in *Suanshu Shijin* into 15 types. Using a reasoning model, we instructed it to output all mathematical methods involved in each problem, resulting in the classification shown in Table 2:

**Table 2** Mathematical Method Classification System and Results

| Mathematical Method Type | Number of Problems | Proportion (%) | Typical Example |
|---|---|---|---|
| Basic Arithmetic Operations | 121 | 14.32% | Land consolidation problems in *The Nine Chapters on the Mathematical Art* |
| Ratio and Fraction Calculations | 262 | 31.01% | Engineering problems in *Zhang Qiujian Suanjing* |
| Area Calculation | 96 | 11.36% | Land area calculations problems in *The Nine Chapters on the Mathematical Art* |
| Volume Calculation | 82 | 9.70% | Granary volume calculations problems in *Sunzi Suanjing* |
| Pythagorean Theorem | 39 | 4.62% | Right triangle problems in *The Nine Chapters on the Mathematical Art* |
| Basic Geometry | 80 | 9.47% | Double difference distance measurement in *The Sea Island Mathematical Classic* |
| Basic Algebra | 21 | 2.49% | Cup-washing problem in *Sunzi Suanjing* |
| Elementary Number Theory | 16 | 1.89% | Unknown quantity problem in *Sunzi Suanjing* |
| Linear Equations or Systems | 67 | 7.93% | Chickens and rabbits count problem in *Sunzi Suanjing* |
| Quadratic Equations | 22 | 2.60% | Square city problem in *Jigu Suanjing* |
| Cubic Equations | 13 | 1.54% | Dimensions of a granary in *Jigu Suanjing* |
| Higher-Degree Equations | 1 | 0.12% | Rectangular and circular storage depth calculation in *Jigu Suanjing* |
| Linear Programming | 1 | 0.12% | Hundred coins chicken problem in *Zhang Qiujian Suanjing* |
| Square Root Extraction | 22 | 2.60% | Area-to-volume conversion in *The Nine Chapters on the Mathematical Art* |
| Arithmetic or Geometric Sequences | 2 | 0.24% | Money distribution problem in *Zhang Qiujian Suanjing* |

Due to the multifaceted nature of some problems, the total number of mathematical methods reached 845 (i.e., an average of 1.57 methods per problem). This highlights the characteristic of ancient Chinese mathematical problems: integration of multiple methodologies. For example, *Zhang*

*Qiujian Suanjing* "Hundred Coins Chicken Problem"(百钱买鸡) involves algebraic equations, optimization theory (linear programming), and integer constraints (number theory), while cubic equation problems in *Jigu Suanjing* often accompany volume calculations. Overall, problems in *Suanjing Shishu* are predominantly elementary mathematics, with knowledge fully covered in Chinese secondary education. Among them, problems requiring basic arithmetic, ratio/fraction calculations, area/volume computations, foundational geometry, Pythagorean theorem, or linear equations dominate.

**2.1.3.2 Difficulty Level Annotation**

Difficulty levels were annotated by referencing the MATH dataset's methodology and adapting it to the mathematical ideas in ancient texts. A four-tier difficulty standard was designed:

Level 1: Most intuitive arithmetic calculations solvable in 1–2 steps using basic operations.

Level 2: Extended applications of arithmetic, including fractions, ratios, Pythagorean theorem, and area/volume calculations.

Level 3: More complex problems requiring advanced techniques (e.g., historical context knowledge, primary equations, or algebra).

Level 4: Highly complex problems demanding integration of multiple mathematical methods, historical knowledge, and potentially geometry, algebra, or number theory, or equations beyond linear/second-degree.

The difficulty criteria were embedded in prompts, instructing the closed-source reasoning model to assign a level to each problem. The final annotations yielded 27, 284, 185, and 42 problems for Levels 1–4, respectively. Table 3 shows the distribution of problems across the eight classics:

**Table 3** Distribution of Problems by Difficulty Level in Ancient Texts

| Classic Name | Level 1 | Level 2 | Level 3 | Level 4 |
|---|---|---|---|---|
| *The Nine Chapters on the Mathematical Art* | 6 | 92 | 88 | 13 |
| *The Sea Island Mathematical Classic* | 0 | 0 | 3 | 6 |
| *Zhang Qiujian Suanjing* | 3 | 35 | 38 | 6 |
| *Xiahouyang Suanjing* | 4 | 60 | 16 | 0 |
| *Wujing Suanshu* | 0 | 5 | 12 | 2 |
| *Jigu Suanjing* | 0 | 0 | 5 | 15 |
| *Wucao Suanjing* | 9 | 51 | 6 | 0 |
| Sunzi *Suanjing* | 5 | 41 | 17 | 0 |

The data in Table 3 reveals two key patterns:

(1) Practical Orientation

The overall difficulty distribution approximates a normal distribution, with low- and high-difficulty problems being less frequent and mid-level problems (Levels 2 and 3) dominating. Level 2 problems, in particular, are most prevalent and concentrated in texts like *The Nine Chapters on the Mathematical Art*, *Xiahouyang Suanjing*, and *Wucao Suanjing*—works closely tied to bureaucratic governance. These problems involve taxation, land measurement, and resource allocation, reflecting the practical focus of ancient Chinese mathematics.

(2) Impact of Textual Research Domains on Difficulty

Texts addressing administrative tasks (e.g., *Wucao Suanjing*) exhibit higher proportions of simple problems (Level 1–2 accounts for 90.9% of its problems). In contrast, texts focused on engineering and measurement (e.g., *Jigu Suanjing* and *The Sea Island Mathematical Classic*) contain more high-difficulty problems (Level 4 constitutes 75% and 66.67% of their problems, respectively).

Fields annotated with problem type and difficulty information were appended to the JSON files described earlier, generating a 7-element tuple for each problem: source, question description, standard answer, solution method, premise conditions, difficulty level, and problem type. This data guided model responses. Figure 4 displays an example tuple from the dataset:

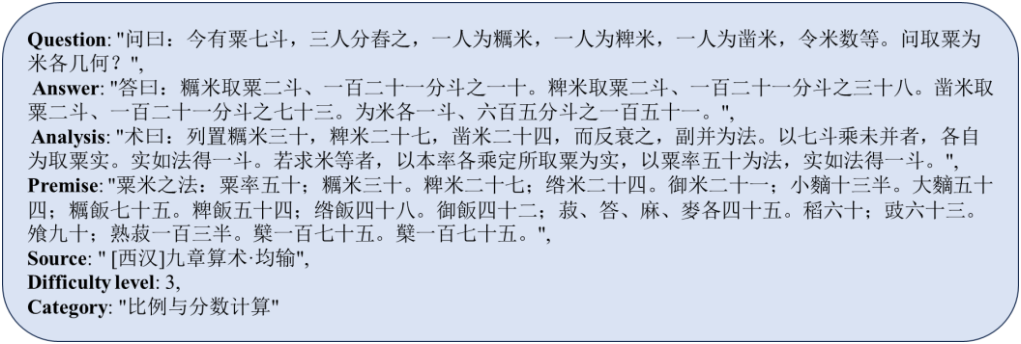

**Question:** "问曰：今有粟七斗，三人分春之，一人为粝米，一人为粺米，一人为凿米，令米数等。问取粟为米各几何？ ",
 **Answer:** "答曰：粝米取粟二斗、一百二十一分斗之一十。粺米取粟二斗、一百二十一分斗之三十八。凿米取粟二斗、一百二十一分斗之七十三。为米各一斗、六百五分斗之一百五十一。 ",
 **Analysis:** "术曰：列置粝米三十，粺米二十七，凿米二十四，而反衰之，副并为法。以七斗乘未并者，各自为取粟实。实如法得一斗。若求米等者，以本率各乘定所取粟为实，以粟率五十为法，实如法得一斗。 ",
 **Premise:** "粟米之法：粟率五十；粝米三十；粺米二十七；凿米二十四。御米二十一；小䴵十三半。大䴵五十四；粝饭七十五。粺饭五十四；糳饭四十八。御饭四十二；菽、荅、麻、麦各四十五。稻六十；豉六十三。飧九十；熟菽一百三半。蘖一百七十五。蘖一百七十五。 ",
 **Source:** " [西汉]九章算术·均输",
 **Difficulty level:** 3,
 **Category:** "比例与分数计算"

**Figure 4.** Example of Evaluation Data

## 2.2 Evaluation Methods and Model Selection

### 2.2.1 Evaluation Methods

For evaluation, we modified prompts and adopted two distinct strategies to assess the performance of open-source and closed-source reasoning models:

(1) **Closed-Book Mode**: Only the problem source, question description, and premise conditions from the JSON tuple were included in the prompt, requiring the model to directly output an answer after reasoning. This mode simulates human problem-solving without external knowledge, focusing on the model's foundational mathematical reasoning capabilities and its ability to interpret classical mathematical concepts.

(2) **Open-Book Mode**: The prompt was expanded to include the problem source, question description, premise conditions, and solution method in ancient math classics from the JSON tuple. The model was instructed to first reference and understand the provided solution in the classical text, replicate all steps of the original solution, and then derive the answer. This mode evaluates the model's ability to transfer ancient methods to modern contexts, particularly its capacity to express traditional algorithmic logic using contemporary mathematical frameworks.

These two modes simulate human approaches to understanding classical mathematical problems and reflect common user practices of leveraging reasoning models to aid comprehension of ancient texts. Employing both strategies effectively measures a model's ability to interpret and solve classical mathematical problems.

### 2.2.2 Model Selection

The open-sourcing of DeepSeek R1 has advanced the democratization of large-scale reasoning models. Recent research has demonstrated that strong long-context reasoning capabilities can be replicated even in mid-sized language models. This study selected several reasoning models with comparable token budgets from recent academic and industrial releases for experimentation. The models are described as follows:

(1) **DeepSeek R1**, a 671B-parameter reasoning model developed by DeepSeek, renowned for robust Chinese comprehension and reasoning.

(2) **Qwen-Plus-2025-04-28**, a closed-source reasoning model by Alibaba Cloud based on the

Qwen3 series (Qwen3/Qwen3_Technical_Report.Pdf at Main · QwenLM/Qwen3, n.d.), achieving superior performance over DeepSeek R1 on benchmarks like AIME and Math500.

(3) **QwQ-32B** (Tongyi Qianwen QwQ-32B, n.d.), a 32B-parameter reasoning model by Alibaba Cloud trained on Qwen2.5 with reinforcement learning focused on mathematics and programming tasks, delivering performance comparable to DeepSeek R1 on specific subtasks.

(4) **Skywork-OR1-32B-Preview** (Skywork-OR1-32B-Preview, n.d.), a reasoning model by Kunlun Tech trained on the Qwen2.5 series.

(5) **DeepSeek-R1-Distill-Qwen-32B**, a model developed by DeepSeek using 800,000 instruction samples to fine-tune the Qwen2.5-32B base model, optimized for general user computational resources.

(6) **Light-R1-32B** (Wen et al., 2025), a 32B-parameter model by 360 Research, the first full-featured reimplementation of DeepSeek-R1 from scratch, trained with two-stage learning and reinforcement learning to acquire reasoning capabilities.

Due to the excessive parameter count of DeepSeek R1, which precluded local deployment, we accessed it and Qwen-Plus-2025-04-28 via API. The remaining four models were deployed locally on two A800 GPUs using the vLLM framework, operating in offline inference mode.

**2.3 Evaluation Metrics and Answer Extraction**

The study uses Accuracy as the metric to evaluate the performance of mathematical problem-solving, calculated as the ratio of the number of correctly answered questions to the total number of questions. For generative models, due to the inherent unpredictability of their outputs, further extraction and processing of the results are required before comparing them with the standard answers to determine correctness. In this experiment, the non-reasoning part of the model's response is first extracted, and then the QwQ-32B model is used to derive the model's answer, which is subsequently compared with the standard answer. The most challenging aspect of this process is unit conversion between the two answers. The measurement units in the answers from the *Suanjing Shishu* are often inconsistent, requiring the establishment of a system based on historical metrological standards. We referenced the conversion relationships between measurement units listed in *Science and Civilisation in China*, combined with all units mentioned in the text, to construct a comprehensive numerical conversion table. This table was then included as part of the input prompt for the QwQ-32B model, instructing the model to first extract the answer and then standardize it to determine whether the model's output could be converted into the numerical value of the original text's answer. The judgment results were used for the final accuracy calculation, thereby quantifying the performance of the reasoning model in solving ancient mathematical problems.

# 3 Experimental Results

## 3.1 Experimental Environment and Hyperparameter Design

The model was executed on a Linux server equipped with two A800 GPUs. The system specifications included CUDA version 12.4, PyTorch framework version 2.5.1, and vLLM version 0.7.2. The following key hyperparameters were selected for model loading:

**Table 4** Hyperparameter Settings for Evaluation Experiments

| Hyperparameter Name | Description | Value |
|---|---|---|
| Max_len | Maximum token length allowed by the model | 32678 |
| temperature | Temperature coefficient controlling randomness (lower values yield more deterministic outputs) | 0.6 |
| top_p | Nucleus sampling probability threshold (retains vocabulary up to cumulative probability) | 0.95 |
| top_k | Sampling candidate word count (only considers top-k words) | 20 |
| repetition_penalty | Repetition penalty factor suppressing redundant content generation | 1 |

These hyperparameter values were informed by guidelines from DeepSeek R1 and Qwen models, which recommend maintaining a temperature coefficient within 0.5–0.7 and using high nucleus sampling thresholds and candidate counts to optimize reasoning performance. Additionally, during deployment, we observed that longer model outputs tend to produce repetitive content. To mitigate this, a higher repetition penalty factor was applied when setting the maximum truncation length to 32678, ensuring output quality and preventing degradation of reasoning effectiveness due to redundancy.

### 3.2 Closed-Book Mode Evaluation Results

The experiment adapted the prompts from the appendix to align with each model's native dialogue template, using a for loop to iterate over JSON tuples and construct complete input prompts. Notably, although all models were trained with reinforcement learning to enclose reasoning processes in paired **<think>…</think>** tags, deployment revealed that some models occasionally bypassed deep reasoning and directly generated answers. To address this, we followed DeepSeek R1's recommendations by appending the </think> token to each prompt to explicitly trigger reasoning capabilities, ensuring every model engaged in some level of analysis before answering. Table 5 summarizes the overall performance of six models in closed-book mode, along with their accuracy across four difficulty levels:

**Table 5** Performance of Reasoning Models on Classical Mathematical Problems in Closed-Book Mode

| Model | Total Correct Answers | Overall Accuracy | Level 1 Accuracy | Level 2 Accuracy | Level 3 Accuracy | Level 4 Accuracy |
|---|---|---|---|---|---|---|
| DeepSeek R1 | **339** | **63.01%** | 77.78% | **65.85%** | **61.08%** | **42.86%** |
| Qwen-Plus-2025-04-28 | 326 | 60.59% | **85.19%** | 63.38% | 57.30% | 40.48% |
| QwQ-32B | 294 | 54.65% | 81.48% | 59.51% | 48.65% | 30.95% |
| Skywork-OR1-32B-Preview | 228 | 42.38% | 62.96% | 44.37% | 39.46% | 28.57% |
| DeepSeek-R1-Distill-Qwen-32B | 152 | 28.25% | 55.56% | 30.28% | 24.32% | 14.29% |
| Light-R1-32B | 236 | 43.87% | 59.26% | 46.83% | 40.54% | 28.57% |

As shown in Table 5, DeepSeek R1 achieved the highest overall accuracy (63.01%), outperforming all 32B-parameter open-source models, particularly in high-difficulty Level 3–4 problems. This highlights the impact of parameter scale on performance. Among 32B models, QwQ-32B demonstrated the best balanced performance, while others lagged by at least 10 percentage points—likely due to superior training data diversity and techniques. DeepSeek-R1-Distill-Qwen-32B performed poorly, as noted in DeepSeek R1's technical report: its limited training (only instruction fine-tuning with minimal mathematical prompts) likely caused overfitting.

Across all models, performance peaked at Level 1 (simple arithmetic), with accuracy declining as difficulty increased, indicating struggles with complex classical problems. Furthermore, all models underperformed compared to their English Math-500 benchmark results (≥70%), despite Math-500 covering only high school-level math. This suggests that models trained on modern Chinese or English struggle to interpret classical mathematical concepts as effectively as they do contemporary language.

To better understand model strengths, we visualized performance across 15 problem types using bar charts.

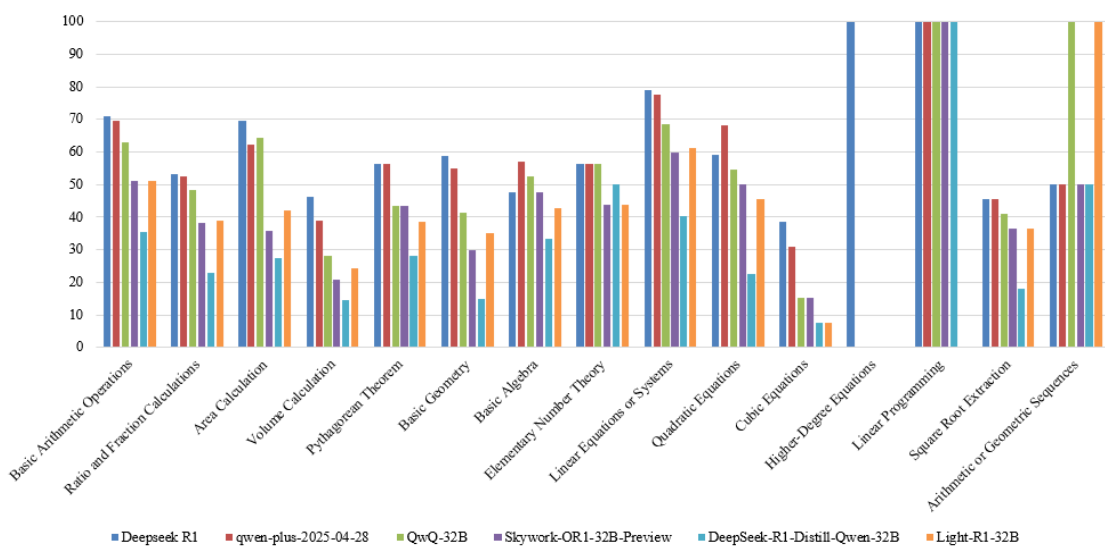

**Figure 5.** Bar Chart of Mathematical Problem-Solving Performance by Reasoning Models in Closed-Book Mode

The above chart illustrates the performance of reasoning models across different mathematical problem types. Excluding problems with extremely low sample sizes (e.g., linear programming, higher-degree equations, and arithmetic/geometric sequences), all models demonstrated strongest performance in basic arithmetic operations, area/volume calculations, and linear equations/systems. Nearly all models achieved their highest accuracy in linear equations/systems, even surpassing performance on simpler arithmetic problems. This may be attributed to the practical relevance of linear equations, which are easier for models to interpret and require fewer complex computational steps, reducing error rates. The two trillion-parameter models (DeepSeek R1 and Qwen-Plus) outperformed smaller-parameter models across nearly all problem types, underscoring their superior understanding capabilities. Smaller models also showed significant performance gaps in challenging problems like quadratic/cubic equations, consistent with Table 5 results.

### 3.3 Open-Book Mode Evaluation Results

In open-book mode, prompts differed from closed-book mode by including solution steps and procedural requirements from the original texts. With a slightly reduced dataset size (adjusted to Level 1: 25, Level 2: 266, Level 3: 178, Level 4: 42), Table 7 summarizes performance in open-book mode:

**Table 6.** Performance of Reasoning Models on Classical Mathematical Problems in Open-Book Mode

| Model | Total Correct | Overall Accuracy | Level 1 Accuracy | Level 2 Accuracy | Level 3 Accuracy | Level 4 Accuracy |
|---|---|---|---|---|---|---|

| | Answers | | | | | |
|---|---|---|---|---|---|---|
| DeepSeek R1 | 354 | 69.28% | 84.00% | 71.05% | 67.42% | **57.14%** |
| Qwen-Plus-2025-04-28 | **359** | **70.25%** | 80.00% | **72.18%** | **70.79%** | 50.00% |
| QwQ-32B | 333 | 65.17% | **96.00%** | 66.17% | 64.04% | 45.24% |
| Skywork-OR1-32B-Preview | 283 | 55.38% | 84.00% | 57.14% | 52.81% | 38.10% |
| DeepSeek-R1-Distill-Qwen-32B | 197 | 38.55% | 72.00% | 39.85% | 35.39% | 23.81% |
| Light-R1-32B | 296 | 57.93% | 76.00% | 62.78% | 52.81% | 38.10% |

In terms of the overall data trend, the distribution pattern of answer accuracy under open-book conditions shows little difference from closed-book mode - models generally excel at solving simple questions while experiencing declining accuracy as problem difficulty increases. Despite a reduction of 23 data points, all models demonstrate significant improvement across nearly all difficulty levels and total correct answers compared to closed-book mode. Light-R1-32B exhibits the most substantial growth in correct responses, solving 60 additional math problems correctly. For all models, Level 4 questions show the greatest performance enhancement, particularly Deepseek R1 and QwQ-32B achieving nearly 15 percentage point improvements. Since the number of Level 4 questions remained unchanged between modes, this indicates that explanatory problem-solving text significantly aids models in addressing complex problems. From model perspective, Qwen-Plus and QwQ-32B demonstrate notable overall performance gains. Qwen-Plus achieves optimal performance, solving 5 more medium-difficulty math problems correctly than Deepseek R1. Meanwhile, QwQ-32B nearly perfectly answers all simple questions, narrowing its performance gap with Deepseek R1 overall. This reflects both the convergence effect of problem-solving methodology provision on model performance enhancement and the particular sensitivity of Qwen series models to medium-difficulty mathematical problems.

We also visualized performance in open-book mode across 15 problem types using bar charts.

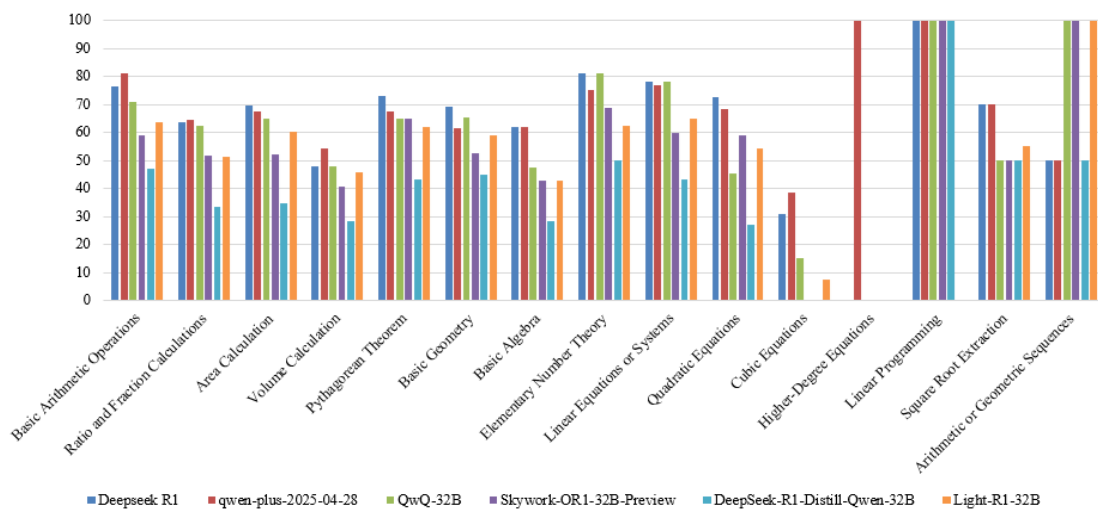

**Figure 6.** Bar Chart of Mathematical Problem-Solving Performance by Reasoning Models in Open-Book Mode

As shown in Figure 6, the inclusion of solution steps led to improved performance across nearly all mathematical problem categories for the models. After excluding problem types with insufficient sample sizes, the models demonstrated significantly higher accuracy in number theory problems and basic arithmetic operations. Notably, the accuracy of elementary number theory problems even surpassed that of linear equations or systems of equations in some cases. This phenomenon may be attributed to the fact that the descriptive solution steps for basic arithmetic and number theory problems more effectively facilitated the models' comprehension of the tasks, whereas the solution methods for linear equations align closely with the models' inherent reasoning patterns, making their inclusion in prompts less influential on the models' cognitive processes.

### 3.4 Analysis of Error Causes in Problem-Solving

The experimental results in the preceding sections demonstrate that accurately answering classical mathematical problems using reasoning models remains a highly challenging task. Compared to the Math-500 dataset, which also restricts problem difficulty to the high school level, even the top-performing models—DeepSeek R1 and Qwen-Plus—achieve only approximately 70% accuracy in open-book mode, significantly lower than their reported performance of over 95% on Math-500. What factors contribute to the poor performance of reasoning models in solving classical mathematical problems? This section provides a quantitative and case-based analysis of the DeepSeek R1 model, which exhibited the most stable and relatively optimal results.

To achieve this, we first invoked the closed-source model's API to compare DeepSeek R1's full reasoning process against the standard solution, requesting the closed-source model to provide a comprehensive analysis of error causes. Subsequently, the reasoning model summarized these errors into nine categories, which were then annotated by both the model and human annotators using a multi-label classification system (allowing multiple error types per problem). The resulting classification framework and annotation results are presented in Table 7:

**Table 7** Classification Framework and Annotation Results for Model Answer Errors

| Category ID | Error Type Description | Annotation Count | Proportion |
|---|---|---|---|
| 1 | Misuse of formulas or models | 56 | 10.94% |
| 2 | Calculation errors | 22 | 4.30% |
| 3 | Format or expression errors | 29 | 5.66% |
| 4 | Lack of specific historical/cultural context | 119 | 23.24% |
| 5 | Misunderstanding of mathematical concepts | 96 | 18.75% |
| 6 | Incorrect problem-solving methods/steps | 42 | 8.20% |
| 7 | Misinterpretation of the problem statement | 40 | 7.81% |
| 8 | Data processing and conversion errors | 105 | 20.51% |
| 9 | Other issues | 3 | 0.59% |

Since the multi-label classification system permits multiple error types per problem, merely counting labels fails to capture intrinsic relationships between error categories. To address this, we further analyzed co-occurrence patterns among error labels, visualizing the results as a heatmap (see Figure 7).

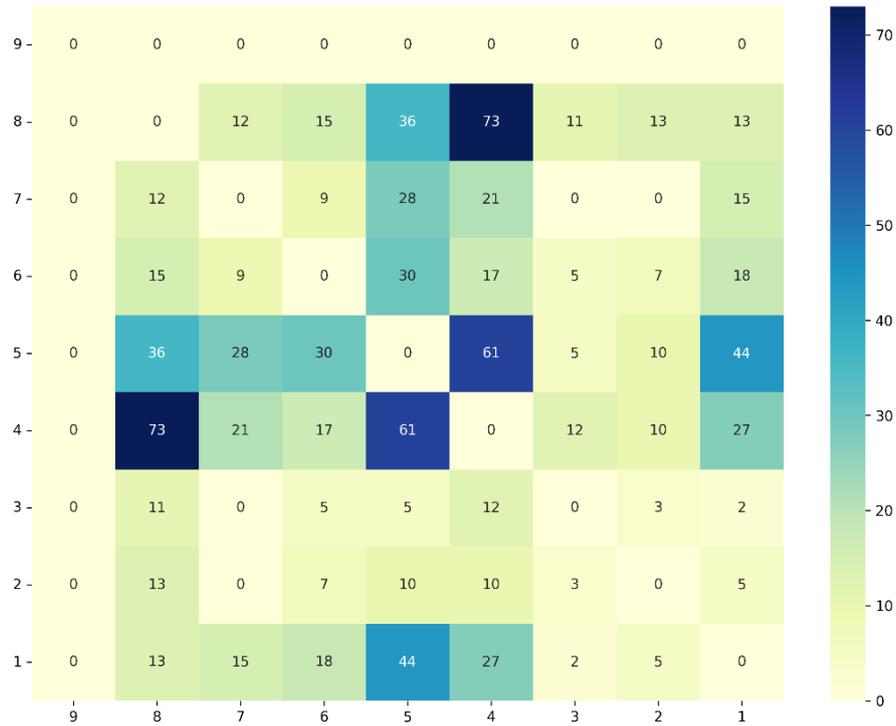

**Figure 7.** Co-occurrence Network Map of Error Labels

As shown in Table 7, the primary sources of errors in the model's mathematical problem-solving are the lack of specific historical and cultural context, misunderstanding of mathematical concepts within the problem, and data processing and conversion errors. Combined with the heatmap analysis, these factors exhibit high co-occurrence frequencies, particularly for label pairs 4–5 (61 co-occurrences) and 4–8 (73 co-occurrences). The inherent characteristics of these labels indicate that when models lack specific cultural context—often tied to unit conversion rules—they frequently misinterpret classical mathematical concepts. The misuse of formulas/models (10.94%) and incorrect problem-solving methods/steps (8.20%), ranking fourth and fifth in error frequency, also show the highest co-occurrence with conceptual misunderstandings in the heatmap. This suggests that the inability to grasp ancient mathematical terminology is a core issue preventing accurate reasoning.

Calculation errors (4.30%) arise not only from computational mistakes in complex operations (e.g., quadratic equations) but also from the model's failure to adhere to ancient mathematical conventions. For instance, ancient Chinese mathematics did not incorporate irrational numbers, yet models sometimes force approximations (e.g., $\pi \approx 3.14$ or $\sqrt{2} \approx 1.414$) into calculations, resulting in discrepancies with original texts. Additionally, 29 problems contained errors in the final result conversion, often due to insufficient understanding of ancient unit conversion rules or difficulties in following traditional numerical notation. In the *Nine Chapters on the Mathematical Art*'s calculation of circular segment area, for example, the model's reasoning might produce a numerically correct value (e.g., 56/81) but incorrectly express it as "五十六分步之八十一" (81/56) instead of the correct "八十一分步之五十六" (56/81), leading to a final answer mismatch.

These findings underscore that misunderstanding mathematical concepts, unit conversion rules, traditional problem-solving logic, and ancient numerical notation are the primary challenges for reasoning models in solving classical Chinese mathematical problems. At their core, these issues stem from a deficiency in specific historical and cultural knowledge. To improve performance on such tasks, models must not only enhance Classical Chinese comprehension but also integrate contextual knowledge about ancient mathematical theories and practices.

# 4 Discussion

This study systematically reveals the performance patterns of mainstream reasoning models in solving classical mathematical problems expressed in Classical Chinese, through the construction of the Guji_MATH benchmark based on the *Suanjing Shishu*. Experimental results indicate that while current reasoning models demonstrate significant capabilities in modern tasks involving mathematical computation or code generation in Mandarin or English, their ability to interpret mathematical problems in Classical Chinese contexts remains notably limited. The following analysis elaborates on the effectiveness and challenges of reasoning models in solving classical mathematical problems from three dimensions: model quality, problem types, and linguistic characteristics.

(1) Model Quality Dimension

In both closed-book and open-book modes, trillion-parameter models (e.g., DeepSeek R1, Qwen-Plus) consistently outperformed 32B-parameter models in overall accuracy. This performance gap widened significantly in high-difficulty Level 3–4 problems. This phenomenon confirms the critical role of parameter scale effects in complex reasoning tasks. Among 32B-parameter models, those trained with higher-quality data and more advanced training stages achieved better accuracy, suggesting that, at equivalent parameter scales, improving data quality and optimizing training algorithms can enhance a model's reasoning generalization to low-resource linguistic tasks.

(2) Problem Type Dimension

Through the construction of difficulty and type labels, this study analyzed reasoning models' performance across diverse mathematical problem categories. Results show that models achieve higher accuracy on simpler problems, such as basic arithmetic operations and linear equations (typically low-difficulty tasks). However, they struggle with more complex modeling and computational challenges, such as quadratic/cubic equations and volume calculations. Future research should focus on optimizing reasoning models to better understand and solve these complex classical mathematical problems.

(3) Linguistic Characteristics Dimension

Experimental results highlight the significant constraints of Classical Chinese texts on model reasoning capabilities, primarily due to comprehension limitations and lack of contextual knowledge. First, Classical Chinese is more concise than Modern Chinese, often omitting mathematical concepts in the original text, making it difficult for models to infer problem intent. Second, classical mathematical problems are not merely computational exercises but also embed complex cultural phenomena and unique ancient scientific systems. For example, the problem of "determining an infant's gender" in the *Sunzi Suanjing* involves China's ancient Yin-Yang theory and numerological traditions, while the calculation of the Moon's declination in the *Jigu Suanjing* is deeply tied to ancient astronomy. Due to insufficient training data, current reasoning models fail to address such problems effectively. Future research must develop models capable of not only mastering Classical Chinese but also deeply understanding Chinese traditional culture and scientific systems to accurately solve ancient mathematical challenges.

# 5 Conclusion and outlook

This study presents the first Guji_MATH benchmark based on *Suanjing Shishu*, systematically evaluating reasoning models' performance in solving mathematical problems expressed in Classical

Chinese. Key findings include:

(1) Model performance exhibits significant scale effects and mode differences. Trillion-parameter models demonstrate clear advantages in handling complex problems, while open-book mode offers smaller models opportunities for localized improvements.

(2) Current reasoning models remain stronger in simpler tasks. Overall problem-solving effectiveness lags behind modern Mandarin or English-based mathematical benchmarks.

(3) Models struggle with deep mathematical integration: Replication of classical algorithms and ancient mathematical reasoning remains superficial, hindered by a lack of cultural context and foundational knowledge.

This research uncovers the coupling mechanism between linguistic structure and mathematical reasoning, breaking through the cultural homogeneity of existing mathematical evaluation frameworks. Methodologically, the proposed "Question-Answer-Method" structured processing framework and dual-mode evaluation system provide transferable solutions for intelligent processing of non-Latin-script classical texts. Results confirm the strong potential of reasoning models in extracting mathematical knowledge from classical texts, with applications in algorithm recreation and aiding comprehension of ancient mathematical works.

However, Our research also have some limitations. Firstly, the dataset is derived solely from the *Ten Mathematical Classics*, covering Han to Tang Dynasty mathematics. Peak achievements of Chinese mathematics during the Song-Yuan-Ming periods (e.g., binomial theorems, trigonometric functions, *Tianyuan Shu* [天元术], *Duoji Shu* [垛积术]) remain unaddressed in the current benchmark. Future work will expand the dataset to evaluate models on these advanced topics. Secondly, problem processing lacks standardization, relying directly on raw text input without exploring whether post-training on mathematical datasets enhances model performance. Subsequent studies will investigate whether translating problems into pseudo-code or training with domain-specific data improves model outputs.

Overall, leveraging reasoning models to solve classical Chinese mathematical problems represents a promising avenue. This approach not only aids modern researchers and enthusiasts in utilizing China's ancient scientific heritage but also facilitates the international dissemination of traditional Chinese mathematical culture.

**Acknowledgements**

The research was funded by the National Social Science Foundation of China (No.21&ZD331).

**Data availability**

Data is applicable.

**Declarations**

**Consent to Publish declaration**

not applicable

**Consent to Participate declaration**

not applicable

**Competing interests**

The authors declare no competing interests.

**Appendix**

**Table 9.** Summary of Model Prompts Used in the Study

| Task Type | Prompt Model | Prompt Example |
|---|---|---|
| Text Punctuation | Xunzi-Qwen1.5-7B_chat | 请为以下文本添加标点符号：{text} |
| Text Structuring | Qwen2.5-14B-Instruct | 接下来将给你一段中国古代数学典籍文本，里面可能涉及到多组数学问题，及其对应的答案和解析，请你使用将每组问题、答案和解析都抽取出来.其中问题往往以'问曰'开头，答案通常以'答曰'开头，解析通常以'术曰'开头，请你按照 json 格式输出多组抽取结果，json 的属性是：问题、答案和解析。如果某项不存在则输出为空\n 给定的文本为:{text} |
| Difficulty Level | qwen-plus-2025-04-28 | 现在将给你一道来自中国古代典籍的数学问题。请你判断这道题目的难度，并在最终给出一个 1-5 分之间的整数作为 |
| Annotation | | 难度评分，数字越大越难，注意由于是古代数学问题，整体难度可能不高，请酌情打分，最终得分用[[]]包裹，比如得[[1]]分，得[[2]]分，你可以参考如下评分等级:1 分题为直观的数学计算，只需适用简单的四则运算可迅速解决\n2 分题多为四则运算的应用，包括简单的求周长和面积等\n3 分题目是更加复杂的应用，需要结合历史知识并使用初次方程等较高级的技术\n4 分题为更复杂的数学问题，可能涉及到几何、代数或数论等领域的综合知识，或者需要多元、多次方程解决\n5 分题为复合性的应用问题，通常同时需要综合历史文化知识、自然科学知识、独特的解题逻辑和多元或多次方程。请根据以上逻辑进行判断。这道题目是: {text} |
| Problem Type Annotation | qwen-plus-2025-04-28 | 现在将给你一道来自中国古代典籍的数学问题。请你给出这道题目所涉及的数学方法，如: 基础四则运算、体积计算、面积计算、几何、初等数论、线性方程组、二次方程、三次方程、高次方程、线性规划、开方计算、比例计算、代数、勾股计算、排列组合等等，你需要使用[[]]包裹最终的判断结果，例如: [[二元一次方程组]], [[线性规划]], [[比例计算]], [[面积计算、三次方程]], [[勾股计算、二次方程、几何]]等(如涉及多项数学技术请使用顿号分开)，这道题目是: {text} |
| Closed-Book Mode Evaluation | The reasoning model presented in Section 2.2.2 | 现在我会向你输入一段来自于古代典籍的数学问题，请你回答该数学问题，在你给出回答时，如果问题涉及到古代计量单位，则你需要结合题目回忆该时代使用的量制单位、度制单位、衡制单位、货币单位、干支历法等类型的计量单位，并将最终的答案需要转化为古代单位的进制，例如: 一百亩又四十二步、二丈八尺三寸四分、十斤十二两三铢、正月乙丑日等等。如果题目不涉及明确的计量单位要求则不需要修改进制，回答可为: 一万五千零四十个，三分之一只，十五天，四分之一天等等。我给你的文本来自{source}，可参考的前提条件是: {premise}，回答以下问题: {question}，请一步一步思考，你的思考过程放在一对<think></think>标签之间，格式如:<think> {用'\n\n' 分隔步骤的思考过程} </think>,在你思考完成后，要将解决方案放在一对<answer></answer>标签之间，格式如: <answer> {最终的、格式化的、精确且清晰的解决方案} </answer>。请你按照上述要求给出思考过程和最终的解决方案。 |
| Open-Book Mode Evaluation | The reasoning model presented in Section 2.2.2 | 现在我会向你输入一段来自于古代典籍的数学问题和对应的解题方法，请你先理解题目和解题方法，并根据解题方法复现原题的解法，解决这道数学问题。在你给出回答时，如果问题涉及到古代计量单位，则你需要结合题目回忆该时代使用的量制单位、度制单位、衡制单位、货币单位、干支历法等类型的计量单位，并将最终的答案需要转化为古代单位的进制，例如: 一百亩又四十二步、二丈八 |

尺三寸四分、十斤十二两三铢、正月乙丑日等等。如果题目不涉及明确的计量单位要求则不需要修改进制，回答可为：一万五千零四十个，三分之一只，十五天、四分之一天等等。我给你的文本来自于{source}，可参考的前提条件是：{premise}，回答以下问题：{question}，本题在原书中的解题方法为：{analysis}，请一步一步思考，你的思考过程放在一对<think></think>标签之间，格式如:<think> {用'\n\n' 分隔步骤的思考过程} </think>,在你思考完成后，要将解决方案放在一对<answer></answer>标签之间，格式如：<answer> {最终的、格式化的、精确且清晰的解决方案} </answer>。请你按照上述要求给出思考过程和最终的解决方案。

| | | |
|---|---|---|
| Answer Consistency Judgment | QwQ-32B | 现在我会向你输入一段涉及古代典籍的数学计算问题，问题的标准答案，以及由某个大语言模型对该问题的输出，你需要从大语言模型的输出中提取出答案文本，然后再和标准答案进行对比，判断是否答对了这个问题。如果答对了判断为正确得 1 分，否则就是错误的得 0 分，问题是：{question}\n 标准答案是：{answer} \n 模型输出是：{pre}\n,请你先从模型的输出文本中提取出最终答案，然后再对比标准答案看是否作答准确，你的回答模板是：问题是 xxx\n 标准的答案是 xxx 从模型输出文本中提取的答案是 xxx\n，xxx(分析文字),所以以是正确/错误的，\n\n 因此最终得[[0]]或[[1]]分。注意得分一定要使用两个中括号括起来。如果不能从模型输出中提取出答案，可以直接认为错误。注意有时模型输出的形式可能与标准答案形式有所差别，如发现经过简单换算或者添加单位后二者相等，或者精确到的单位差异很小时，则同样认为正确，但计算错误时则不可认为正确。另外，请注意古文中一些特殊的数字表述，比如假设存在五位数，第四位为 0 时，则可以忽略。如一万五百等于一万零五百，不可认为这二者不相等。太半、少半分别代表 2/3 和 1/3。在转换时，可以参考如下单位换算：1 |

丝 ＝10 忽,\
,1 毫 ＝10 丝\
,1 氂 ＝10 毫\
,1 分 ＝10 氂\
,1 寸 ＝10 分\
,1 尺 ＝10 寸\
,1 丈 ＝10 尺\
,1 引 ＝10 丈\
,1 匹/疋 ＝4 丈\
,1 端 ＝5 丈\
,1 步 ＝6 尺\
,1 顷 ＝100 亩\
,1 亩 ＝240 步\

,1 里 = 300 步\
,1 圭 = 10 粟\
,1 撮 = 10 圭\
,1 抄 = 10 撮\
,1 勺 = 10 抄\
,1 合 = 10 勺\
,1 升 = 10 合\
,1 斗 = 10 升\
,1 斛 = 10 斗\
,1 絫 = 10 黍\
,1 銖 = 10 絫\
,1 两 = 24 銖\
,1 斤 = 16 两\
,1 钧 = 30 斤\
,1 石 = 4 钧\
,1 贯 = 1000 文。

| Error Cause Analysis | Deepseek R1 | 现在我会向你输入一段来自于古代典籍的数学问题,真实答案,以及由模型输出的思考过程和答案,但模型对这道题目的最终回答却是错误的,现在你需要分析为什么模型做错了这道问题。我给你的文本来自于 {source},这道题目对应的前提条件是:{premise},这道题目为:{question},这道题目的正确答案是:{answer},本题在原书中的解题方法为:{analysis},模型的解题思路和结果是:{pre},请一步一步思考分析模型的思考过程和错误的原因,你的思考过程放在一对<think></think>标签之间,格式如:<think> {用'\n\n' 分隔步骤的思考过程} </think>,在你思考完成后,要将错误原因的分析放在<answer></answer>标签之间,格式如: <answer> {错误原因分析:xxx\n 可将错误原因总结为:[[xxx]](中括号中为简短的错误归因)} </answer>。请你按照上述要求给出错误分析过程和模型回答的错误原因。 |